\documentclass[runningheads]{llncs}
\usepackage[T2A]{fontenc}  
\usepackage{graphicx}
\usepackage{amsmath}
\usepackage{amssymb}
\usepackage{comment}
\usepackage{wrapfig}
\usepackage[english]{babel}
\usepackage{float}
\usepackage{hyperref}
\usepackage[ruled, linesnumbered]{algorithm2e}
\usepackage[strings]{underscore}

\DontPrintSemicolon
\renewcommand{\u}[1]{{\upshape#1}}

\newcommand{\var}[1]{\texttt{#1}}

\newcommand{\deltapos}{\Delta^+}
\newcommand{\deltaneg}{\Delta^-}

\begin{document}
\title{\texorpdfstring{%
Delta Schema Network\\%
in Model-based Reinforcement Learning%
}{%
Delta Schema Network\\%
in Model-based Reinforcement Learning%
}}
%
%
\author{Andrey Gorodetskiy\inst{1, 2} \and
Alexandra Shlychkova\inst{1} \and
Aleksandr I. Panov\inst{1, 3}}
\authorrunning{A. Gorodetskiy et al.}
%
\institute{Moscow Institute of Physics and Technology (National Research University), Moscow, Russia
\and 
Bauman Moscow State Technical University, Moscow, Russia
\and
Artificial Intelligence Research Institute, Federal Research Center ``Computer Science and Control'' of the Russian Academy of Sciences, Moscow, Russia
\email{gorodetskiyandrew@gmail.com}}
\maketitle   
\begin{abstract}
This work is devoted to unresolved problems of Artificial General Intelligence - the inefficiency of transfer learning. One of the mechanisms that are used to solve this problem in the area of reinforcement learning is a model-based approach. In the paper we are expanding the schema networks method which allows to extract the logical relationships between objects and actions from the environment data. We present algorithms for training a Delta Schema Network (DSN), predicting future states of the environment and planning actions that will lead to positive reward. DSN shows strong performance of transfer learning on the classic Atari game environment.
\keywords{Reinforcement learning \and Model-based \and Schema Network \and Delta Schema Network \and Transfer Learning}
\end{abstract}

\section{Introduction}
For an intelligent agent acting in real-world conditions, it is necessary to generalize the experience gained in order not to learn from scratch after a slight change in the environment. A human does not relearn the policy of interaction with a familiar object, but only slightly corrects it, when object's characteristics are changed. For this, logical relationships between objects and their characteristics are used at different levels of generalization. For example, in the Atari game \textit{Breakout}, the colors of the bricks do not matter and the natural agent does not change the policy when colors change. Artificial agent can achieve such a generalization using some universal model-based learning algorithm. In this paper, we propose a new approach to the learning of universal models for reinforcement learning in game environments - Delta Schema Network (DSN) - which is an extension of the early work Schema Network \cite{kansky2017schema}. 

A Schema Network is an object-oriented model, the main aspect of which is a schema. In this architecture the agent receives an image from the environment, which is parsed into a set of extracted objects. Then model learns a set of rules~- schemas, which reflect logical interconnection between objects' properties, actions and rewards. Each schema predicts some property of the objects of certain type, using information about properties of other objects and actions from past observations.

It is possible to represent this interconnection as a factor graph. A variable node in this graph is either a property of some object, potentially achievable reward or action. Factors are schemas, that have input and output nodes. The edges indicate the presence of a causal relationship between the objects and events. An agent can find a node with a positive reward in future time layers of this graph and plan actions to reach it.

Since the graph has fairly generalized properties, the trained Schema Network can be used in conjunction with some feature extractor on environments with similar interaction dynamics. This provides advantage in transfer learning. However, planning on a sufficiently large graph for a long time horizon can be a very challenging task for real-world applications.

From the point of view of creating AGI systems, the DSN algorithm can be used to automatically generate scripts for the behavior of an intelligent agent. These scripts can be used to speed up the agent's own behavior planning process \cite{Panov2016b,Kiselev2019}, or to predict user behavior in a cognitive assistant scenario \cite{Smirnov2019b}.

\section{Related work}

Representation of logical interconnection often helps to increase an efficiency of transfer learning. Various methods are used to represent the logical relationships of objects: in the Schema Network \cite{kansky2017schema}, these are specially introduced schemas with binary logic. In Logical Tensor Networks \cite{serafini2016logic}, it is proposed to use real logic. To further apply the obtained relationships for planning, one can use them as additional data for a neural network. For example, in \cite{toyer2018action} schemas are passed to a neural network. Authors in \cite{badreddine2019injecting} add logical relationships to the input of a neural network using logical tensor network. Another approach is to build a dependency graph and search for a reachable state with a positive reward.

The usage of the Schema Network in reinforcement learning consists of two main stages: training a network to predict future states of environment and construct a prediction graph with the subsequent search for the best reachable reward node.

The Schema Network uses object-oriented approach described in \cite{diuk2008object}. During the training stage, due to the knowledge of the types of environment objects, a model is able to identify the logical connections between them. A similar approach was used in Interaction Network \cite{battaglia2016interaction}, for which, however, no planning algorithms were developed to obtain a positive reward. For similar problems convolutional neural networks are used as in \cite{kipf2016semi}. However, this approach requires a prior knowledge about the structure of the graph, while the Schema Network allows to obtain knowledge about relations between objects automatically from the environment.

Also, during Schema Network training stage a dependency graph is constructed. Finding the reachable state of the environment in which a reward is received can be considered as a estimate of the posterior maximum and solved using the max-product belief propagation \cite{attias2003planning}.

\section{Model description}

\subsection{Main concepts}

Key concepts used in the Delta Schema Network (DSN) are entities, attributes and schemas. Entity is any object that can be extracted from the image. Attribute is a binary variable that indicates presence or absence of a specific property of an entity, each entity has the same $M$ number of attributes. Attribute with value of $1$ or $True$ is said to be \textit{active}.
Schema is a logical AND function that predicts value of attribute or reward at time step $t$, taking as arguments arbitrary number $k$ of attributes and actions at previous time steps $\{t^*: 0 < t - t^* < d\}$.
\[
Schema\colon (Attributes_{t^*} \cup Actions_{t^*})^k \to Attributes_{t} \cup Rewards_{t}
\]
Schemas are represented as binary column vectors and forms parameter matrices. We define $W = (W_{i}: i = 1 .. M)$ to be a tuple of parameter matrices used for attribute prediction, one matrix per attribute type. Parameter matrix used for reward prediction we denote as R.

DSN model learns dynamics of the environment in terms of schema vectors and, from some point of view, represents both transition and reward functions of the environment.
Using learned vectors, model predicts next states of the environment and plan a sequence of actions that will lead to reward.

\subsection{State representation}

In our work we considered each pixel of image as entity. However, we think this model is more suitable for reasoning on more high-level concepts. Attributes of entities have meaning of presence in this pixel object of a certain type, i.e entity's attribute vector is one-hot encoded type of this entity concatenated with void attribute that can indicate absence of any object in this pixel.

DSN relies on semantic information about observation from environment, namely which type of object each pixel belongs to. As observation at time step $t$ model gets state matrix $s_t$ of $(N, M)$ shape, where $N$ is the number of entities an $M$ is the number of attributes. This matrix is suggested to be built from image of $N$ pixels and $M-1$ object types. We consider $s_t$ is provided by some feature extractor.

\subsection{Prediction}

Schema vectors are used to predict \textit{changes} (deltas) in the current state $s_t$ and to predict reward $r_t$ after taking action $a_t$.
If there are several schemas that predict same attribute, their results are united using logical OR.
Two types of schemas are used for attribute prediction: creating, represented by $W^+$, and destroying, represented by $W^-$. Creating schemas predict attributes that are not active in $s_t$, but should be active in $s_{t+1}$. Vice versa, destroying schemas predict destruction of attributes that are active in $s_t$, but should disappear in $s_{t+1}$.

We used $d = 2$, i.e. for computing attribute value at $t+1$ schema can use attributes and actions only at $t$ and $t-1$. Thus, frame stack has size $2$.

To make a prediction on either attributes or rewards we construct \textit{augmented matrix} $X_t$, which is built from the frame stack $(s_{t-1}, s_t)$ and action $a_t$ in the following way:
\begin{enumerate}
\item $s_{t-1}$ and $s_t$ are augmented into $s_{t-1}^*$ and $s_t^*$, correspondingly. Each row, which is an attribute vector of some entity, is horizontally concatenated with attribute vectors of these entity's $R-1$ spatial neighbors. Referring to corresponding image, these neighbors are located in square around the central pixel, and the central pixel is represented by a row in $s$.
\item $a^*$ is built, which is broadcasted by number of rows to $s_t$ version of one-hot encoded action $a_t$.
\item horizontally concatenated  $(s_{t-1}, s_t, a^*)$ result in $x_t$.
\end{enumerate}

To predict next state $s_{t+1}$, we predict creating $\deltapos$ and destroying $\deltaneg$ state changes using $W$ and then apply them to $s_t$:
\[
\deltapos_j = \overline{\overline{X}_t W_j^+} \vec{1} \qquad\qquad
\deltaneg_j = \overline{\overline{X}_t W_j^-} \vec{1}\,,
\]
where $\Delta_j$ denotes $j$th column of $\Delta$.
\[
s_{t+1} = s_t - \deltaneg + \deltapos \,,
\]
considering elements as integers and clipping result after every operation in chain to $\{0, 1\}$.

Reward is predicted in similar way:
\[
r_{t+1} = \vec{1}^{\intercal} \, \overline{\overline{X}_t R} \, \vec{1}
\]
This matrix multiplication of augmented matrix and parameter matrix is equivalent to applying discrete convolutions to the original image, where parts of images are described by rows of augmented matrix and filters are column vectors in parameter matrix. Thus, DSN models the environment as cellular automaton - grid of entities - and reconstructs its rules as schema vectors.

\section{Learning algorithm}

During interaction with the environment agent stores unique transitions in replay buffer.
We use learning algorithm from \cite{kansky2017schema} with different target in a self-supervised manner. First, correctness of already learned schema vectors is checked on new observations. Schema vectors that produce false positive predictions are deleted.
After that, we learn new schema vectors. Targets for learning $W$ are columns of $\deltapos$ and $\deltaneg$, which we denote as $y$. Target for learning $R$ is the reward, obtained at sample's time step.

\begin{enumerate}
\item We choose one random sample with false negative prediction from replay buffer and put it in the set $solved$.

\item We solve the following LP optimization problem: finding a schema vector $w$ that does not produce any false positive predictions on replay buffer, predicts positive labels for all samples in $solved$ and maximizes the number of true positive predictions on replay buffer.
\begin{align*}
\min_{w\in\{0,1\}^{D}} & \sum_{n: y_n=1} (1-x_n)w\\
\text{s.t. } & (1-x_n)w > 1 \quad \forall_{n: y_n=0}\\
& (1-x_n)w = 0 \quad \forall_{n \in \text{solved}}\\
\end{align*}

\item All samples that got predicted by obtained schema vector are added to the set $solved$.

\item We try to simplify schema vector: minimizing its $L_1$ norm with condition of absence false positive predictions on replay buffer and false negative on set $solved$.
\begin{align*}
\min_{w\in \{0,1\}^{D}} & w^T\vec{1}\\
\text{s.t. } & (1-x_n)w > 1 \quad \forall_{n: y_n=0}\\
& (1-x_n)w = 0 \quad \forall_{n \in \text{solved}}\\
\end{align*}
\end{enumerate}

\section{Planning algorithm}

The purpose of planning is to find a sequence of actions that will lead to positive reward.
The planning process consists of several stages:
\begin{enumerate}
    \item Forward pass builds factor graph of potentially reachable nodes;
    \item A set of target reward nodes are selected;
    \item Sequence of actions that will activate target node are planned.
\end{enumerate}

The input to the planner is the frame stack of size $2$ consisting of state matrices
$(s_{t-1}, s_t)$ and schema parameters $(W^+, W^-, R)$.

\subsection{Forward pass}

The future states of the environment are predicted for $T$ time steps ahead.
Simultaneously, we build the factor graph $G$ in which variable nodes are attributes, rewards or actions; factors are schemas that were activated during prediction process and edges connect schemas to their input and output nodes. Every node has assigned time step, at which it appeared in prediction. Thus, graph $G$ is said to be consisting of layers, that unite nodes within same time step.

To predict next state one need to decide which action $a$ agent takes at current state. When predicting $\deltapos$, DSN model assumes that agent takes all possible actions, and for $\deltaneg$ it assumes agent takes ''do not do anything`` action. This leads to superimposing of all possible $\deltapos$ for the next state in the single matrix $s_{t+1}$.

During forward pass we maintain graph building in the following way.
After predicting $s_{t+1}$, for each predicted attribute or reward node at layer $t+1$ we instantiate on the graph concrete instances of corresponding \textit{creating} schemas.
Each attribute at $s_t$ that was not destroyed by any of the \textit{destroying} schemas is considered to be active at $s_{t+1}$ and we mark the corresponding node at $t+1$ as having self-transition, that is like a schema with single input that activates attribute at time $t+1$ provided it was active at $t$.

\subsection{Target nodes selection}

Having predictions for the future states of the environment on $T$ ticks ahead, reward nodes of the factor graph are added to the target queue.

\medskip
\begin{tabular}{rl}
$q$ &= sorted by time potentially reachable positive reward nodes\\
& $= [r^+_{closest} \dots r^+_{farthest}]$
\end{tabular}
\medskip

\subsection{Finding sequence of actions}

We take next reward node from queue $q$ and try to find a sequence of actions for agent to reach it.
To find such sequence, one need to find a configuration of graph $G$, that satisfy following constraints:
\begin{itemize}
\item target reward node is active
\item at each layer $t$ only \textit{one} action node is active
\end{itemize}

In this configuration, the values of the attribute and reward nodes show their actual reachability.
The action nodes $\{a_i \in G: i \in [1..T]\}$ represent the actions that must be taken to reach the target node.

Some of the learned schema vectors may depend on actions, while in the dynamics of the environment there is no such dependence.
This occurs because during training events correlated, but did not have a causal relationship.
For correct planning, it is necessary to find a valid configuration of the graph, constructed by predictions with such vectors.
We propose the \var{backtrace_node} algorithm (algo. \ref{algo:backtrace_node}) to find such a configuration. It does not perform exhaustive search, but works well in our experiments.

We maintain an array of joint constraints on the active action nodes for each time layer. During graph traversal, we either satisfy these constraints or replan paths to nodes committed to these constraints if there is no other path to the target. Process of node activation goes in the following order:
\begin{itemize}
\item try to activate the node by self-transition
\item try to activate the node with an action-independent schema
\item if there is no constraint on the current tick, try to activate the node with any schema
\item try to select a schema that satisfies the constraint on the current tick
\item replan all vertices that require current constraint
\begin{itemize}
    \item[\textbullet] find a set of actions that as constraints would allow the activation of each conflicting node
    \item[\textbullet] sequentially start replanning subgraph of each conflicting node using the action acceptable by all
    \item[\textbullet] if all nodes have been replanned successfully, change the constraints at conflicting layer
\end{itemize}
\end{itemize}

During the replanning process, a new conflict situation may arise. Then new replanning process should be recursively started.

\section{Experiments}

\begin{wrapfigure}{r}{0pt}
    \centering
    \includegraphics[width=0.2\textwidth]{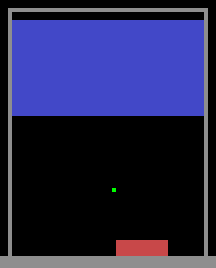}
    \caption{Breakout}
    \label{fig:ataribreakout}
    \vspace{-\baselineskip}
\end{wrapfigure}

The model was evaluated on the Atari Breakout game (see Figure~\ref{fig:ataribreakout}). The goal of the game is to knock down bricks with a ball, substituting a moving platform under it. There are no random factors in the environment.

The action space consists of the following actions: {do not move, move left, move right}.
As an observation, the agent receives an RGB image and information about a particular type of object each pixel belongs to. Rewards are distributed as follows: {+1 for knocking down a brick, -1 for dropping a ball past the platform, 0 in other cases}.

\begin{figure}[htbp]
    \centering
    \includegraphics[width=.9\textwidth]{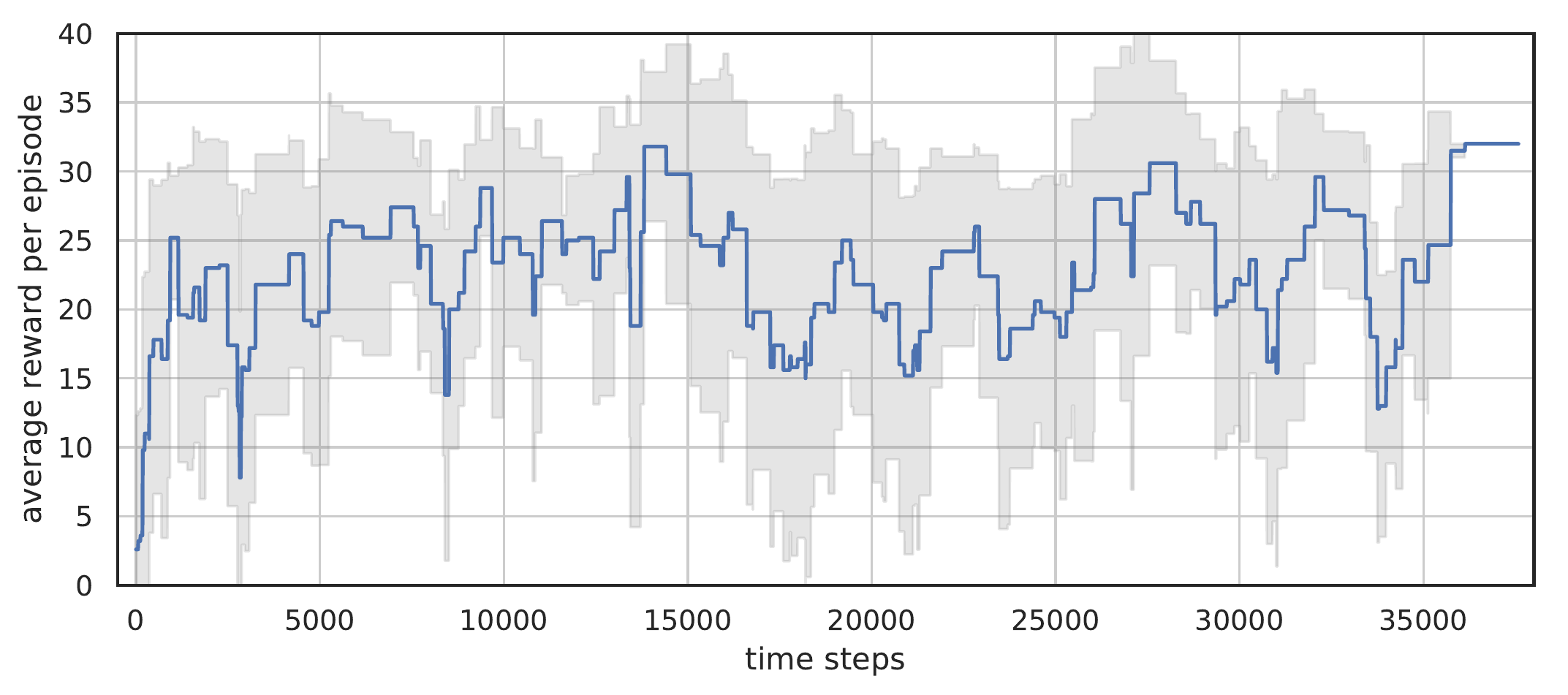}
    \caption{Total reward per episode of DSN on standard Breakout.
    Averaged over 5 runs, shaded region represents the standard deviation.}
    \label{fig:experiment-dsn}
\end{figure}

The number of schema vectors for each parameter matrix was limited to $500$ units.
The episode was limited to 5000 steps, agent had 3 lives after the loss of which the episode ended. Highest possible reward for episode was 36. Figure \ref{fig:experiment-dsn} shows the results of DSN evaluation.

Agent did not managed to knock down completely all bricks in part of episodes, because after destroying some part of them to the top wall, it could not longer detect future reward and hence plan actions.

\begin{figure}[htbp]
    \centering
    \includegraphics[width=.9\textwidth]{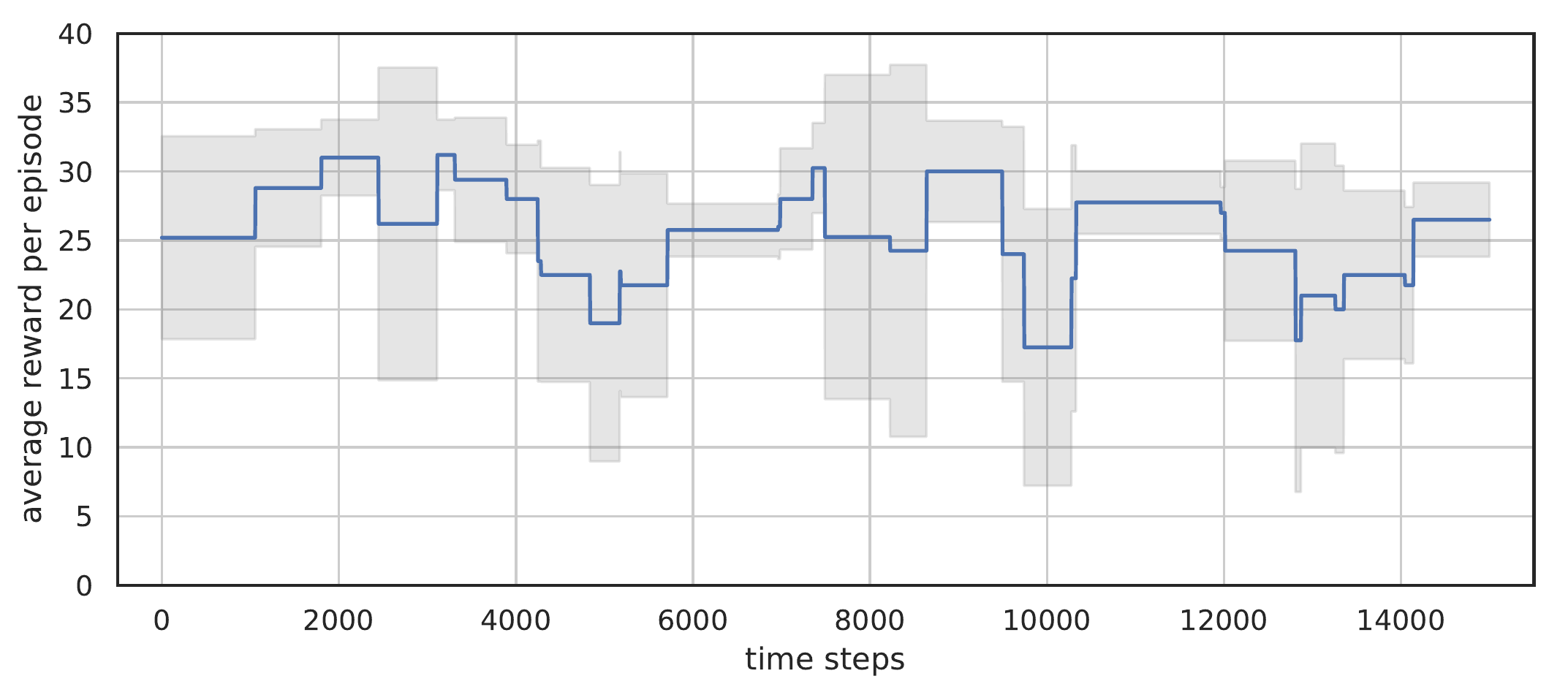}
    \caption{DSN performance after zero-shot transfer. Averaged over 5 runs, shaded region represents the standard deviation.}
    \label{fig:experiment-transfer}
\end{figure}

A distinctive feature of the DSN is the efficient transfer of the trained model to environments with similar dynamics.
We evaluated the model, trained in the previous experiment, on the same environment but with two balls.
Results of transfer without additional training (see Figure~\ref{fig:experiment-transfer}) show similar average score.

\begin{figure}[htbp]
    \centering
    \includegraphics[width=0.95\textwidth]{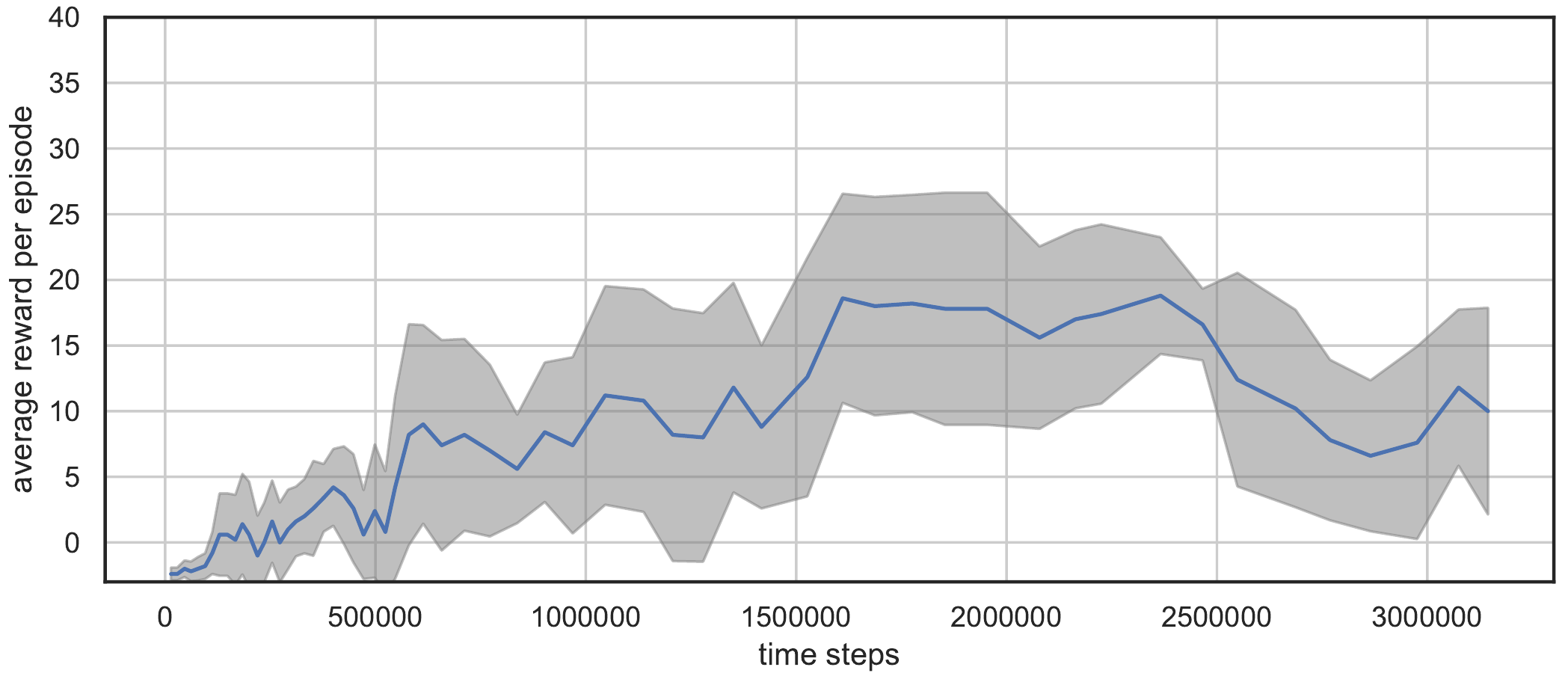}
    \caption{DQN training process on standard Breakout. Single run, rolling mean with window size = 5. Shaded region represents standard deviation of window samples.}
    \label{fig:experiment-dqn}
\end{figure}

We compared DSN model to DQN. Figure \ref{fig:experiment-dqn} shows that DQN needs significantly more time steps to reach equal performance.

\section{Conclusion}
In this paper, we proposed an original implementation of the universal logical model of environment dynamics for model-based reinforcement learning. Our approach, which we called Delta Schema Network, is a modification and extension of Schema Network for RL. We described in detail the algorithmic implementation of the proposed method and conducted basic experimental studies on the Atari Breakout environment. 

Future works include the use of logical model-based approaches for real-world robotic tasks, such as controlling a robotic manipulator \cite{Mnih2015HumanlevelCT,Younes2019} or a car at an road intersection \cite{Shikunov2019}. Code of the DSN model can be obtained in the repository: \href{https://github.com/cog-isa/schema-rl}{github.com/cog-isa/schema-rl}.

\textbf{Acknowledgements.} The reported study was partially supported by RFBR, research Projects No. 17-29-07079 and No. 18-29-22027.

\bibliographystyle{splncs04}
\bibliography{references}

\clearpage
\section{Appendix}

Algorithms \ref{algo:backtrace_node_by_schemas} and \ref{algo:backtrace_schema} are used in algorithm \ref{algo:backtrace_node}.
\renewcommand{\labelitemi}{\textbullet}
Node in the graph is considered to have next attributes:
\begin{itemize}
\item node.is_reachable - the actual reachability of the node, subject to currently selected actions, or \var{None} if the reachability is not known.
\item node.schemas - map from actions to node's schemas requiring these actions
\item node.transition - self-transition node, if any
\end{itemize}
\begin{algorithm}[htb]
\SetKwInOut{Input}{Input}
\SetKwInOut{Output}{Output}
\Input{node - target node \\
       schemas - set of available schemas}
\Output{actual node reachability, planned actions}
\BlankLine
\For{\u{schema in schemas}}{
    backtrace_schema(schema)\;
    \uIf{\u{schema.is_reachable}}{
        node.is_reachable${}\gets{}$True\;
        break\;
    }
}
\caption{backtrace_node_by_schemas}
\label{algo:backtrace_node_by_schemas}
\end{algorithm}
\begin{algorithm}[htb]
\SetKwInOut{Input}{Input}
\SetKwInOut{Output}{Output}
\Input{schema - target schema\\
       preconditions - input nodes of schema}
\Output{actual schema reachability}
\BlankLine
schema.is_reachable${}\gets{}$True\;
\For{\u{precondition in preconditions}}{
    \uIf{\u{precondition.is_reachable} is None}{
        backtrace_node(precondition)\;
    }
    \uIf{not \u{precondition.is_reachable}}{
       schema.is_reachable${}\gets{}$False\;
        break\;
    }
}
\caption{backtrace_schema}
\label{algo:backtrace_schema}
\end{algorithm}
\begin{algorithm}[htb]
\SetKwInOut{Input}{Input}
\SetKwInOut{Output}{Output}

\Input{
\renewcommand{\labelitemi}{\textbullet}
\begin{itemize}
\item node - target node to backtrace
\item desired_constraint=None - desired constraint at node.t${}-1$ to satisfy, if any
\end{itemize}
}
\Output{
actual node reachability, planned actions in joint_constraints
}
\BlankLine
node.is_reachable${} \gets False$\;
\uIf{\u{desired_constraint} is not None}{
    backtrace_node_by_schemas(node, schemas[desired_constraint])\;
    \Return
}
\tcp{try to activate the node by self-transition}
\uIf{\u{node.transition.is_reachable} is None}{
    backtrace_node(node.transition)\;
}
node.is_reachable ${}\gets{}$ node.transition.is_reachable\;
\lIf{\u{node.is_reachable}}{\Return}
\tcp{try to activate the node with an action-independent schema}
backtrace_node_by_schemas(node, schemas[All action independent])\;
\lIf{\u{node.is_reachable}}{\Return}
\uIf{\u{no current constraint}}{
    backtrace_node_by_schemas(node, schemas[All action dependent])\;
    \uIf{\u{node.is_reachable}}{
    set new constraint for current layer in joint_constraints}
    \Return
}
\tcp{try to select a schema that satisfies the current constraint}
backtrace_node_by_schemas(node, schemas[current_constraint])\;
\uIf{\u{node.is_reachable}}{
    add current node as committed to current constraint\;
    \Return
}
\tcp{replan all vertices that require current constraint}
negotiated_actions${}\gets {}$actions acceptable by all conflicting nodes\;
is_success = False\;
\For{\u{action in negotiated_actions}}{
    backtrace_node(node, desired_constraint=action)\;
    \uIf{\u{node.is_reachable}}{
        is_success = True\;
        \For{\u{curr_node in committed_nodes}}{
            backtrace_node(curr_node, desired_constraint=action)\;
            \uIf{not \u{curr_node.is_reachable}}{
                curr_node.is_reachable = True\;
                is_success = False\;
                break
            }
        }
    }
    \uIf{\u{is_success}}{
    change constraints for current layer to new ones\;
    break\;
    }
}
\caption{backtrace_node(node, desired_constraint=None)}
\label{algo:backtrace_node}
\end{algorithm}

\end{document}